%% file: 03-arxiv.tex
\documentclass[runningheads]{llncs}

\usepackage{eccv}

\usepackage{eccvabbrv}

\usepackage{graphicx}
\usepackage{booktabs}
\usepackage[autostyle]{csquotes}

\usepackage[accsupp]{axessibility}  % Improves PDF readability for those with disabilities.

\usepackage[pagebackref,breaklinks,colorlinks,citecolor=eccvblue]{hyperref}
\usepackage{orcidlink}

\input{preamble}

\begin{document}

\title{ProMerge: Prompt and Merge\\
for Unsupervised Instance Segmentation}

% \titlerunning{ProMerge}

\author{Dylan Li\inst{1,*}\orcidlink{0009-0004-3507-3442} \and
Gyungin Shin\inst{2,*}\textsuperscript{\Letter}\orcidlink{0000-0003-1793-665X}}

\authorrunning{D.~Li et al.}

\institute{Meta Reality Labs \and
Visual Geometry Group, University of Oxford\\
\url{https://www.robots.ox.ac.uk/~vgg/research/promerge}}

\maketitle
\blfootnote{
\textsuperscript{*}Equal contribution\\
\textsuperscript{\Letter}Correspondence to: gyungin@robots.ox.ac.uk}

%%%%% Main %%%%%
\input{sections/00-abstract}
\input{sections/01-introduction-post-rebuttal}
\input{sections/02-related-work-post-rebuttal}
\input{sections/03-method}
\input{sections/04-experiments-post-rebuttal}

\input{sections/05-discussion}
\input{sections/06-conclusion}
\input{sections/acknowledgements}
%%%%%%%%%%

%%%%% Supplementary Materials %%%%%
\appendix
\chapter*{Appendix}

\input{supplementary-sections/introduction}
\input{supplementary-sections/pseudo-code}
\input{supplementary-sections/further-implementation-details}
\input{supplementary-sections/further-ablation-studies}
\input{supplementary-sections/further-visualisations}

%%%%%%%%%%

\bibliographystyle{splncs04}
\bibliography{main}
\end{document}

%% file: preamble.tex
\usepackage{array}
\usepackage{amsmath}
\usepackage{amssymb}
\usepackage{commath}
\usepackage[toc,page]{appendix}
\usepackage{lmodern}
\usepackage{multirow}
\usepackage{multicol}

\UseRawInputEncoding

\usepackage[capitalize]{cleveref}
\crefname{section}{Sec.}{Secs.}
\Crefname{section}{Section}{Sections}
\Crefname{table}{Table}{Tables}
\crefname{table}{Tab.}{Tabs.}

\usepackage{enumitem}
\usepackage{graphicx}
\usepackage{pifont}% http://ctan.org/pkg/pifont
\usepackage{xspace}
\usepackage{xcolor}
\usepackage{colortbl}
\usepackage{algorithm}
\usepackage{algpseudocode}
\usepackage{tabularx}
\usepackage{hyphenat}  % to force linebreak a long hyphenated noun phrase
\usepackage{marvosym}
\definecolor{noel_green}{RGB}{61,137,37}
\definecolor{noel_grey}{RGB}{179,179,179}
\definecolor{green_cell}{RGB}{29,177,0}
\definecolor{orange_cell}{RGB}{255,147,0}
\definecolor{grey_cell}{RGB}{230,231,231}
\definecolor{algo_green}{RGB}{65,128,126}
\usepackage[hang,flushmargin]{footmisc}

\newcommand{\comment}[1]{{\textcolor{algo_green}{#1}}}

\newcommand{\mypara}[1]{\vspace{2pt}\noindent{\bf{#1}}}
\newcommand{\qcr}[1]{{\fontfamily{qcr}\selectfont{#1}}}
\newcommand{\xmark}{\ding{55}}%

\newcommand{\methodName}{ProMerge\xspace}

\newcommand{\simpleap}[0]{\text{AP}}

\newcommand{\maskap}[0]{\text{AP}^{\text{mk}}}
\newcommand{\maskapfifty}[0]{\text{AP}^{\text{mk}}_{50}}

\newcommand{\maskar}[0]{\text{AR}^{\text{mk}}_{100}}

\newcommand{\boxap}[0]{\text{AP}^{\text{bb}}}
\newcommand{\boxapfifty}[0]{\text{AP}^{\text{bb}}_{50}}

\newcommand{\boxar}[0]{\text{AR}^{\text{bb}}_{100}}

\newlength\savewidth\newcommand\shline{\noalign{\global\savewidth\arrayrulewidth
  \global\arrayrulewidth 1pt}\hline\noalign{\global\arrayrulewidth\savewidth}}

\newcolumntype{H}{>{\setbox0=\hbox\bgroup}c<{\egroup}@{}}

\newcommand\blfootnote[1]{%
  \begingroup
  \renewcommand\thefootnote{}\footnote{#1}%
  \addtocounter{footnote}{-1}%
  \endgroup
}

%% file: sections/00-abstract.tex
\begin{abstract}
Unsupervised instance segmentation aims to segment distinct object instances in an image without relying on human-labeled data. This field has recently seen significant advancements, partly due to the strong local correspondences afforded by rich visual feature representations from self-supervised models (e.g., DINO).
Recent state-of-the-art approaches use self\hyp{}supervised features to represent images as graphs and solve a generalized eigenvalue system (i.e., normalized\hyp{}cut) to generate foreground masks.
While effective, this strategy is limited by its attendant computational demands, leading to slow inference speeds.
In this paper, we propose Prompt and Merge (\methodName), which leverages self-supervised visual features to obtain initial groupings of patches and applies a strategic merging to these segments, aided by a sophisticated background-based mask pruning technique.
ProMerge not only yields competitive results but also offers a significant reduction in inference time compared to state\hyp{}of\hyp{}the\hyp{}art normalized\hyp{}cut\hyp{}based approaches. Furthermore, when training an object detector using our mask predictions as pseudo-labels, the resulting detector surpasses the current leading unsupervised model on various challenging instance segmentation benchmarks.

\mypara{Keywords:} Unsupervised Instance Segmentation $\cdot$ Prompt and Merge
\end{abstract}

%% file: sections/01-introduction-post-rebuttal.tex
\section{Introduction}
Instance segmentation identifies and delineates each distinct object within an image, providing both its class and precise pixel-wise location. This capability is crucial for a wide range of applications, from autonomous driving systems~\cite{cordts2016cvpr} that must navigate complex environments to medical imaging technologies that require accurate tumor segmentation~\cite{bilic2019liver, he2023samsurvey}.
However, the cost of manually annotating dense masks for training data is prohibitively high, especially for domains such as medical imaging that require deep expertise.

\input{figures/teaser}

To overcome the challenges with dense annotations,
multiple endeavors have attempted to tackle category-agnostic instance segmentation in an unsupervised manner,\footnote{In this paper, we consider the \textit{class-agnostic} setting following the prior works~\cite{wang2022tokencut,wang2023cvpr} on unsupervised instance segmentation.} among which normalized\hyp{}cut\hyp{}based approaches~\cite{wang2022tokencut, wang2023cvpr} have recently shown promise.
These methods partition a graph representation of an image encoded in feature space into similar parts and solve a generalized eigenvalue system, specifically through spectral clustering with normalized\hyp{}cut~\cite{shi2000tpami}.
Notably, the recently introduced MaskCut method~\cite{wang2023cvpr}, which iteratively employs the TokenCut algorithm~\cite{wang2022tokencut} on a single image multiple times to generate numerous instance masks, demonstrates state\hyp{}of\hyp{}the\hyp{}art performance. However, repeatedly solving this generalized eigenvalue problem incurs significant computational demands that delay the per-image inference time.
Furthermore, its reliance on a fixed criterion to determine the number of segmentation masks per image (with MaskCut using three) may not capture the complete taxonomy of objects in dense, intricate scenes.

In our paper, we propose a simple yet effective framework called \methodName, which sidesteps the aforementioned limitations. 
We start with generating initial masks of locally grouped patches by point-prompting self-supervised visual features (e.g., DINO~\cite{caron2021dino}).
The initial masks, generated through computing the affinity between individual local patches and global patches, constitute a large set of mask proposals covering different parts of a given image.
Following this mask generation process, we iteratively merge these local masks based both on their pixel overlap and their similarity in feature space. 
The effectiveness of \methodName is demonstrated through its faster inference speed (about 3.6 times) and competitive performance compared to existing training-free\footnote{Here, training-free is defined as not requiring training for segmentation.} unsupervised methods on six benchmarks, including the densely annotated LVIS~\cite{gupta2019cvpr} and SA-1B~\cite{kirillov2023iccv} datasets.
Moreover, by training an object detector (i.e., Cascade Mask R-CNN~\cite{cai2018cvpr}) with the mask predictions by \methodName as pseudo-labels, we show that our framework outperforms the current leading unsupervised model (i.e., CutLER~\cite{wang2023cvpr}).

Our contributions are three-fold:
(i) We propose the \methodName framework, composed of an initial mask generation step using point-prompted self-supervised visual features, an iterative mask merging that considers similarities in pixel and feature spaces, and a sophisticated mask pruning strategy;
(ii) We compare the competitive performance of our approach to the popular normalized\hyp{}cut\hyp{}based methods on six standard instance segmentation benchmarks, including COCO2017~\cite{lin2014eccv}, COCO-20K~\cite{vo2020eccv}, LVIS~\cite{gupta2019cvpr}, KITTI~\cite{geiger2012cvpr}, and subsets of Objects365~\cite{shao2019iccv} and SA-1B~\cite{kirillov2023iccv};
(iii) When training a class-agnostic object detector using the predictions by \methodName as pseudo-labels, we show the resulting detector outperforms the leading unsupervised detector on the above datasets. 

%% file: figures/teaser.tex
\begin{figure}[!t]
    \centering
    \includegraphics[width=.99\textwidth]{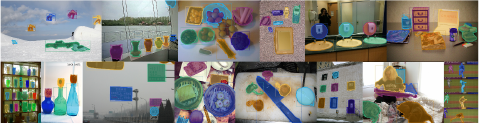}
    \caption{\textbf{Qualitative examples of \methodName, a simple yet effective \textit{training-free} approach for unsupervised instance segmentation.} Despite its simplicity, \methodName demonstrates strong segmentation performance.}
    \label{fig:teaser}
\end{figure}

%% file: sections/02-related-work-post-rebuttal.tex
\section{Related work}
Our work is connected to three themes in the literature, including self-supervised visual representation learning, unsupervised single object detection/segmentation and unsupervised instance segmentation.

\subsection{Self-supervised visual representation learning}
Self-supervised learning in computer vision has advanced significantly by adopting the principle of learning from the intrinsic structure of data, drawing inspiration from how language models such as GPT~\cite{brown2020neurips} and BERT~\cite{devlin2019bert} achieve semantic understanding from text.
One strategy in self-supervised learning involves leveraging pretext tasks, such as those employed by Masked Autoencoders~\cite{MaskedAutoencoders2021}, wherein models learn by predicting the obscured parts of an image. Another set of strategies, embodied by SwAV~\cite{caron2020unsupervised}, MoCo~\cite{he2019moco, chen2020mocov2, chen2021mocov3}, and DINO~\cite{caron2021dino, oquab2023dinov2, darcet2023vitneedreg}, uses data augmentations to generate varied perspectives of images and aligns feature representations of these perspectives.
Among these self-supervised paradigms for training encoders, DINO in particular encodes detailed segmentation information, a capability diminished in models trained with supervised labels ~\cite{caron2021dino,shin2022cvprw}.
In this work, by leveraging DINO's inherent grouping ability, we propose a simple yet effective approach to instance segmentation without explicit labels.

\subsection{Unsupervised single object detection and segmentation}
Unsupervised object detection and segmentation aim to localize a single, dominant object in an image in an unsupervised manner with a bounding box or a segmentation mask, respectively. 

LOST ~\cite{oriane2021bmvc} extracts features from a self-supervised network and isolates the foreground by first identifying the seed patch with the lowest count of positive correlations with other patches. A seed expansion strategy is then used to include additional patches that correlate positively with the original seed patches.

Another line of work uses normalized\hyp{}cut-based methods~\cite{wang2022tokencut,melas2022cvpr,shin2022cvprw} to distinguish the foreground from the background. This class of methods uses the eigendecomposition of the Laplacian matrix derived from a feature affinity matrix constructed with self-supervised features. The resulting eigenvectors, processed with traditional clustering or thresholding methods, can be translated into meaningful segmentations in pixel space. 

This process facilitates the identification of coherent regions, enabling the separation of primary image elements from the background.
While these prior works serve as foundational methods for unsupervised multi-object localization and segmentation, they primarily focus on segmenting or identifying the location of a \textit{single} salient object, limiting their generalizability and performance on multi-object localization tasks.
\subsection{Unsupervised instance segmentation}
Unsupervised instance segmentation aims to identify \textit{multiple} objects in an image without human labels, introducing a more complex challenge than the aforementioned single object detection and segmentation. 
An early attempt~\cite{vangansbeke2022arxiv} generates a single salient mask per image, and trains the Mask R-CNN detector~\cite{he2017iccv} using the generated masks as pseudo-labels.
However, the single mask generation approach does not provide sufficient mask instances per image, resulting in suboptimal detection performance.

Recent approaches~\cite{wang2022cvpr,wang2023cvpr}, on the other hand, propose methods for generating multiple instance masks per image, that are used to train a general object detector.
MaskCut~\cite{wang2023cvpr} has demonstrated promising outcomes by iteratively applying a normalized\hyp{}cut\hyp{}based single object segmentation technique (i.e., TokenCut~\cite{wang2022tokencut}) to images and training a robust object detector with pseudo-labels generated through the MaskCut algorithm. Despite its effectiveness, MaskCut uses repeated eigenvalue system resolutions for the normalized cut, incurring significant computational demands. Additionally, MaskCut limits itself to three segmentations per image.
In contrast, ProMerge eschews computationally intensive eigenvalue calculations in favor of using raw feature affinities and does not impose a restriction on the number of segmentations per image.
By doing so, it offers a more computationally efficient alternative for the instance segmentation task while achieving higher precision and recall.

%% file: sections/03-method.tex
\section{Method}\label{sec:method}
In this section, we first describe the problem scenario (\cref{subsec:problem-formulation}) and introduce \methodName, a simple yet effective prompt-and-merge method to tackle unsupervised instance segmentation (\cref{subsec:promerge}).
We then describe a background-based mask pruning strategy to extract foreground masks from a set of prompted masks that increases the effectiveness of prompting and merging (\cref{subsec:mask-pruning}). We finally describe a pseudo-label training scheme in which an object detector is trained with pseudo-labels generated by \methodName (\cref{subsec:self-training}).

\input{figures/overview}
\subsection{Problem formulation}\label{subsec:problem-formulation}
We address the challenging task of instance segmentation in an unsupervised manner. Given an image $\mathbf{I} \in \mathbb{R}^{3 \times H \times W}$, where $H$ and $W$ denote the height and width of the image, we aim to produce a set of $N$ instance binary masks $\mathcal{M} = \{\mathbf{M}_l \in \{0,1\}^{H \times W}| l={1,...,N}\}$ without relying on any human-labeled data.

\subsection{ProMerge: Prompt and Merge}\label{subsec:promerge}

\mypara{Point-prompting visual features. }
Our approach begins with generating preliminary mask proposals. We use visual features, denoted as $\mathbf{F} = \{\mathbf{f}_{ij} \in \mathbb{R}^{c}| i=1,...,h,~j=1,...,w \}$, that are obtained by feeding an image into an image encoder (e.g., patch tokens for Vision Transformers (ViT)~\cite{dosovitskiy2021iclr}). Here, $c$, $h$, and $w$ represent the channel, height, and width dimensions of the features.
To generate initial masks from the visual features (i.e., patch tokens), we consider the technique of point-prompting, wherein a 2D grid of $K$ equally spaced patch tokens is selected as seeds for mask generation.
This subset of the selected tokens $\mathcal{P}=\{\mathbf{p}_l \in \mathbb{R}^{c}|l=1, ..., K\}$, which we call the set of \textit{prompt tokens}, is individually compared with all of the patch tokens in the image. By comparing each prompt token in a one-to-all manner via a similarity measure, we generate an affinity matrix $\mathbf{A}_l \in [-1, 1]^{h \times w}$ for each prompt token $\mathbf{p}_l$.
In this paper, we use \textit{key} features from the last attention layer of a self-supervised image encoder (i.e., DINO) as patch tokens~\cite{wang2023cvpr} and compute the cosine similarity between them.
That is,
\begin{equation}
\mathbf{A}_l = (A_{l;ij}) = \frac{\mathbf{p}_l \cdot \mathbf{f}_{ij}}{||\mathbf{p}_l||_2||\mathbf{f}_{ij}||_2}
\end{equation}
where $||.||_2$ denotes L2 norm. We then apply a bipartition threshold, $\tau_b$, to the affinity matrix to obtain a binary mask $\mathbf{M}_l$.

\mypara{Merging prompted masks.}
Given the prompted masks above, we consider an iterative clustering method, wherein masks, sorted by area in descending order, are sequentially merged with a set of larger masks processed at the previous iterations.
The processed masks serve as bases for merging smaller masks that are introduced in later iterations.

In comparing a new, smaller mask with a previously processed, larger mask, we consider two straightforward conditions.
First, we use the Intersection-over-Area (IoA) metric to determine if the smaller mask should be merged with the larger mask.
If the ratio of the intersection area between two masks to the area of the smaller mask exceeds a certain threshold, denoted as $\tau^\text{merge}_{\text{IoA}}$, the smaller mask is combined with the larger.
We choose IoA over the more conventional Intersection-over-Union (IoU) due to IoU's diminished effectiveness in cases where a large mask merges with a significantly smaller one, such as when combining an object's main body with an appendage.
Note that the IoA criterion alone is insufficient, as it prevents merging two masks unless they exhibit substantial overlap in pixel space.
To allow for merging semantically similar masks with a small overlap, we consider the second condition based on feature similarity. Given a pair of masks, we compute the cosine similarity between their average patch embeddings and merge them if the feature similarity is over a threshold $\tau^\text{merge}_{f}$. 
If a new mask meets at least one of the merging conditions with more than one previously processed mask, the new mask and all compatible masks are merged together. Conversely, if it does not satisfy a merge condition with any of the previously processed masks, it remains unchanged. This mask will then be compared with subsequent masks in later iterations for potential merging.

In summary, through our merge algorithm, we start with disparate, potentially overlapping masks, and end up with semantically-related mask groupings.

\subsection{Background-based mask pruning}\label{subsec:mask-pruning}
While the prompt-and-merge framework thus far is intuitive, we observe that it suffers from poor performance in isolation, due to the noisy background masks among the prompted masks.
We introduce a mask pruning strategy based on background prediction between the prompting and merging steps. We rely on a two-step process that (i) groups the initial $K$ masks into the foreground or background, and aggregates the background masks via pixel-wise voting to produce a single, fine background mask for the image and (ii) uses the predicted background mask to filter out noisy foreground masks from the initial mask proposals. Each step in the mask pruning strategy is detailed below.

\mypara{Background aggregation.}
Recall that after prompting the visual features with $K$ equally spaced points in a 2D grid, we obtain $K$ binary masks.
We then classify each of the masks as either a foreground or background candidate.
We employ a simple heuristic: a background mask is likely to contain a majority of pixels along the edge for at least two of the edges of the image. Specifically, we consider a mask as background if more than one of its sides contains a number of positive pixels that exceeds half the length of that side.
Then, we create a single representative background mask for the image by applying a pixel-wise voting scheme to the background candidates.
Formally, given the set of background candidate masks $\mathcal{B} = \{\mathbf{M}^\text{bg}_l\}$, a pixel value at $(i, j)$ of the aggregated background mask $\tilde{M}^\text{bg}_{ij}$ is determined with:

\begin{equation}
    \tilde{M}^\text{bg}_{ij} = \Biggl[ \frac{\sum_{l=1}^{|\mathcal{B}|} M^\text{bg}_{l;ij}}{|\mathcal{B}|} > 0.5 \Biggl]
\end{equation}
where the $[.]$ operator is the indicator function, which returns 1 if the condition within the operator is satisfied, and 0 otherwise.
The condition in the operator checks whether over the half of the background candidate masks have a value of 1 at $(i,j)$.
As such, we obtain a single representative background mask per image, represented by $\tilde{\mathbf{M}}^\text{bg} \in \{0,1\}^{h \times w}$. 
Some visual examples are shown in \cref{fig:pixelwise-voting}.

\input{figures/pixelwise-voting}

\mypara{Foreground filtering.}
With the background mask $\tilde{\mathbf{M}}^\text{bg}$, we exclude prompted masks that are more likely to belong to the background before the merge process.
For this filtering step, we consider three separate approaches: (i) intersection-based (ii) similarity-based filtering and (iii) the proposed Cascade filtering.

For intersection-based filtering, we use a simple prior that any foreground masks considered for the following merge process should not significantly intersect with the voted background $\tilde{\mathbf{M}}^\text{bg}$. Similarly to \cref{subsec:promerge}, we use IoA, as we want the metric to be invariant to the size disparity between a candidate foreground mask and $\tilde{\mathbf{M}}^\text{bg}$.
If the intersection of a foreground mask and $\tilde{\mathbf{M}}^\text{bg}$ divided by the area of the mask is greater than $\tau_\text{IoA}^\text{bg}$, we regard the mask as belonging to the background and exclude it in the merging process. 

For similarity-based filtering, we
prune masks with a high similarity with the background mask in feature space.
We again compute the mean patch embeddings for the candidate foreground  mask and $\tilde{\mathbf{M}}^\text{bg}$, before calculating their normalized cosine similarity. 
If the similarity value is over a threshold, we exclude the mask from the merging process.

\input{figures/cascade-filtering}

In the proposed Cascade filtering approach (shown in \cref{fig:cascade-filtering}), we initially sort the prompted masks in ascending order based on their area. We then proceed through these masks sequentially, maintaining a cumulative ledger of pixels already incorporated into previous masks.
At each step, we identify pixels in the current mask that have not yet been considered in a prior mask.
We then calculate the IoA between these `new' pixels and $\tilde{\mathbf{M}}^\text{bg}$.
We also compute the feature similarity between the current mask and $\tilde{\mathbf{M}}^\text{bg}$ in a manner similar to the aforementioned similarity-based filtering.
If either of these measures with $\tilde{\mathbf{M}}^\text{bg}$ exceeds their respective thresholds, the mask is excluded from the subsequent merge process.

\subsection{ProMerge+: Training an object detector with pseudo-labels from \methodName}
\label{subsec:self-training}
Following~\cite{wang2023cvpr}, we train an object detector on \methodName predictions generated from inference on images in a large-scale image dataset (i.e., ImageNet2012~\cite{deng2009cvpr}).
The purpose of this pseudo-label training is two-fold: firstly, to obtain an object detector with better performance by learning from the noisy pseudo-labels;
and secondly, to assess the detector's ability to generalize across different data distributions by training on images from one dataset (e.g., ImageNet2012) and evaluating on another (e.g., SA-1B~\cite{kirillov2023iccv}).
The trained detector, ProMerge+, surpasses performance and zero-shot transfer capabilities of the current leading model.

%% file: figures/overview.tex
\begin{figure}[!t]
    \centering
    \includegraphics[width=.98\textwidth]{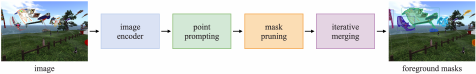}
    \caption{\textbf{An overview of \methodName.} Given an input image, we obtain initial mask proposals by prompting visual features from an image encoder using a 2D point grid. Then, the noisy proposals are filtered through the proposed background-based mask pruning. The final predictions are made by iteratively merging the remaining foreground masks.}
    \label{fig:overview}
\end{figure}

%% file: figures/pixelwise-voting.tex
\begin{figure}[!t]
    \centering
    \includegraphics[width=.98\textwidth]{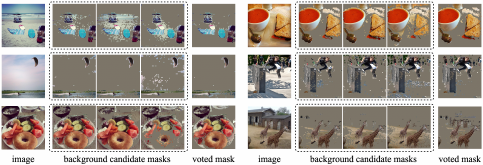}
    \caption{\textbf{Qualitative examples of the pixel-wise voting.} For each case, an input image, background candidate masks (only three masks are shown for visual purposes), and the voted mask are visualized. The voted background mask, $\Tilde{\mathbf{M}}^\text{bg}$ effectively filters out the background, leaving only foreground regions despite the noisy candidate masks.}
    \label{fig:pixelwise-voting}
\end{figure}

%% file: figures/cascade-filtering.tex
\begin{figure}[!t]
    \centering
    \includegraphics[width=.98\textwidth]{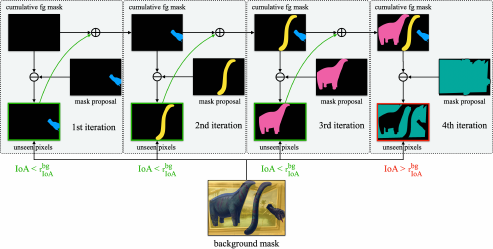}
    \caption{\textbf{An illustration of the proposed Cascade mask filtering process.}
    For each iteration of the proposed method, we evaluate the newly proposed mask by focusing on the pixels that have not yet been covered by the cumulative foreground mask, which is an aggregation of pixels from mask proposals in preceding iterations. If these previously unseen pixels demonstrate a significant overlap with the background mask, quantified by the Intersection-over-Area (IoA) metric, the mask proposal for that iteration is subsequently disregarded. An example of this can be observed in the fourth iteration (rightmost), in which the mask proposal is eliminated due to its high IoA with the background mask. Note that in the figure, the feature similarity condition is not shown for visual clarity. See the text for details.}
    \label{fig:cascade-filtering}
\end{figure}

%% file: sections/04-experiments-post-rebuttal.tex
\section{Experiments}\label{sec:experiments}
In this section, we first provide implementation details (\cref{subsec:implementation-details}) and compare our method with the state-of-the-art-methods (\cref{subsec:main-results}). Then, we provide extensive ablation study to analyze our approach (\cref{subsec:ablation-study}).
\subsection{Implementation details}\label{subsec:implementation-details}
Our implementation is based on the PyTorch~\cite{adam2019neurips} and Detectron2~\cite{wu2019detectron2} libraries, and A100 GPUs are used for our experiments unless otherwise stated.

\mypara{Datasets.}
We evaluate our \methodName on six benchmarks including COCO2017~\cite{lin2014eccv}, COCO-20K~\cite{vo2020eccv}, LVIS~\cite{gupta2019cvpr}, KITTI~\cite{geiger2012cvpr}, subsets of Objects365~\cite{shao2019iccv} and SA-1B~\cite{kirillov2023iccv} (44K and 11K images, respectively).
Among these, SA-1B is the most challenging benchmark due to the densely-annotated fine-grained masks, with an average of 101 segmentation masks per image.
To train ProMerge+, we use unlabeled ImageNet2012 training images~\cite{deng2009cvpr} (1.2M images) and evaluate the resulting model on the six benchmarks in a zero-shot manner (i.e., the model is not trained with images sharing the same data distribution as evaluation data).
For the ablation study, we use COCO2017 following~\cite{wang2023cvpr}.

\mypara{Evaluation metrics.}
We evaluate our methods based on average precision (AP) and recall (AR), the standard metrics for the instance segmentation task.

\mypara{Inference of \methodName.}
We follow the previous work~\cite{wang2023cvpr} for our inference setting. Specifically, we use DINO~\cite{caron2021dino} with the ViT-B/8 architecture~\cite{dosovitskiy2021iclr} as an image encoder and input images are resized to 480$\times$480 pixels before being fed into the encoder. We apply Conditional Random Field (CRF)~\cite{crf2011neurips} to an output mask for post-processing.
Additionally, we split connected components of initial masks before merging, which we find beneficial in terms of both AP and AR (shown in \cref{subsec:ablation-study}).

\mypara{Pseudo-label training.}
When we train an object detector with pseudo-masks generated by \methodName, we use the Cascade Mask R-CNN model~\cite{cai2018cvpr} with the ResNet50 backbone~\cite{he2016cvpr} initialized with DINO features~\cite{caron2021dino}.
We train the detector for 160K iterations using the SGD optimizer with a batch size of 16, a momentum of 0.9, and a learning rate of 0.005, which is decreased by a factor of 5 during training.

\input{tables/comparison-sota}

\subsection{Main results}\label{subsec:main-results}
In this section, we compare our methods to the state-of-the-art methods with both standard evaluation metrics and inference speed.

\mypara{Comparison to state-of-the-art methods.}
We first compare \methodName to \textit{training}-\textit{free} methods including TokenCut~\cite{wang2022tokencut} and MaskCut~\cite{wang2023cvpr} algorithms.
In \cref{tab:sota} (top), \methodName demonstrates superior performance in both average precision ($\simpleap$) and average recall ($\text{AR}_{100}$) compared to the existing methods.
The overall higher recall of \methodName is attributed to its flexibility in not requiring a predetermined number of masks per image.
Qualitative results of each method can be found in the supplementary.

\input{tables/comparison-speed}

We next compare ProMerge+, a class-agnostic object detector trained with the predictions of \methodName on  ImageNet as pseudo-labels, to the state-of-the-art model (i.e., CutLER~\cite{wang2023cvpr}).
As can be seen in \cref{tab:sota} (bottom), ProMerge+ outperforms CutLER~\cite{wang2023cvpr} by 0.9 and 1.4 in $\simpleap$ and $\text{AR}_{100}$, respectively, on average.
Note that we use the same detector architecture (i.e., Cascade Mask R-CNN~\cite{cai2018cvpr}) and the training recipe as CutLER. The sole difference is that CutLER is trained with predictions from MaskCut and ProMerge+ is trained with ones from \methodName.
These results demonstrate the effectiveness of using our higher quality pseudo-labels for training an object detector.

\mypara{Speed comparison.}
Unlike the state-of-the-art training-free methods that depend on solving the computationally intensive normalized-cut algorithm, our method avoids this complexity, enabling faster inference. To demonstrate this, we compare the inference speed of our method with that of normalized\hyp{}cut\hyp{}based methods in terms of Frames Per Second (FPS).
In detail, we measure the timing for each of TokenCut, MaskCut, and \methodName across 200 images, using the first 100 images as a warm-up phase to ensure accurate measurements.\footnote{We use a single RTX 3080 GPU and a 12th Gen Intel(R) Core(TM) i7-12700K chipset for this experiment.}
As shown in \cref{tab:speed-comparison}, \methodName runs at 0.54 FPS, which is about 1.6 and 3.6 times faster than TokenCut and MaskCut, respectively.
In addition, we compare the speed differential between ProMerge and MaskCut under various inference configurations. For MaskCut, we adjust the number of repetitions ($t$) for solving the eigenvalue problem, which significantly impacts its runtime. For our method, we vary the number of prompt tokens ($K$).
As can be seen in \cref{fig:stride-speed}, 
ProMerge provides a more advantageous trade-off between inference speed and performance in $\maskap$ compared to MaskCut.
It achieves higher $\maskap$ values when $K$ exceeds 100, while also being significantly faster---approximately 4.9 times faster at $K=100$.

\input{tables/effect-of-each-component}

\subsection{Ablation study}\label{subsec:ablation-study}
Here, we conduct an extensive ablation study to analyze the effect of each component in \methodName. Specifically, we explore the effects of pixel-wise voting for background aggregation, foreground filtering methods, different merging conditions, and varied hyperparameter choices.

\mypara{Effect of each component.}
We identify major components that affect the performance of our approach:
(i) prompting and merging;
(ii) background-based mask pruning (denoted as Mask Pruning); and
(iii) connected component splitting (denoted as CC Splitting).
In \cref{tab:each-component}, we show the influence of each component.
Notably, a naive approach that relies solely on
prompting and merging suffers from poor performance, with 0.4 $\maskap$ and 1.0 $\maskar$, which are worse than a prompting-only approach.
However, adding background\hyp{}based mask pruning greatly boosts both $\maskap$ and $\maskar$ by 1.7\% and 3.7\%. CC-splitting further increases $\maskap$ by 0.3\% and $\maskar$ by 2.8\%.
These results demonstrate that though the core of prompt-and-merge is straightforward, the competitive performance of our approach is facilitated by incorporating sophisticated components such as background mask pruning.

\mypara{Effect of pixel-wise voting.}
In the background-based mask pruning process, we use a pixel-wise voting strategy to aggregate background candidate masks to produce a single, representative background mask for a given image.
We consider the case where we substitute voting with a simpler non-voting mechanism, and compare it with the voting based mechanism in \methodName. In the non-voting experiment, we sum up all background candidate masks. If a pixel at location $(i, j)$ has at least one mask with a value of one (i.e., positive pixel), it is regarded as a background pixel. As highlighted in \cref{tab:effect-of-voting}, voting leads to a significant enhancement in both average precision and recall.
In the absence of pixel-wise voting, the aggregated background often encompasses not only the actual background but also parts of the foreground in the image, which detracts from the overall performance.
Conversely, the application of the voting strategy more accurately isolates the background region.

\input{tables/effect-of-voting}

\mypara{Effect of foreground filtering method.}
Given the background mask obtained from the pixel-wise voting above, we filter prompted masks based on the background mask as described in \cref{subsec:mask-pruning}.
Here, we compare three different methods including (i) IoA-based, (ii) feature-similarity-based, and (iii) our proposed Cascade filtering strategies.
For the IoA-based filtering, we discard a mask proposal if it has an IoA with the background mask exceeding 0.5, meaning more than half of the proposal's pixels are part of the background.
In the feature-similarity-based approach, we classify a proposal as background if the cosine similarity between the normalized mean embeddings of the candidate foreground mask and the background exceeds 0, implying that the proposal's embedding closely aligns with that of the background.

From \cref{tab:effect-of-filtering-conditions}, we make the following key observations.
First, IoA-based filtering, demonstrates higher recall but lags in precision compared with similarity-based filtering.
This discrepancy can be attributed to the latter's tendency to filter out more masks, irrespective of their degree of area overlap with the background, which in turn affects recall.
Second, our proposed Cascade filtering emerges as the superior method.
Its effectiveness stems from retaining smaller object masks while filtering larger masks that might incorrectly merge these smaller ones (in the following merge step). In other words, as Cascade filtering focuses specifically on the new, unseen pixels (and their respective relationships with the background), it lifts precision and recall by preventing the amalgamation of smaller entities into a larger, encompassing mask during mask merging.

\mypara{Effect of merging conditions.}
In the merging process, we consider two different conditions for merging: (i) Intersection-over-Area (denoted as IoA) and (ii) feature similarity (denoted as feat.) between two masks.
As shown in \cref{tab:effect-of-merging-conditions}, we observe that the feature similarity condition plays a more significant role than the IoA condition when used separately.
However, allowing both conditions leads to the best performance, indicating that they are complementary conditions.

\input{tables/effect-of-filtering-condition}

\mypara{Hyperparameter analysis.}
Lastly, we conduct experiments to explore the impact of various hyperparameters, including grid size, bipartition threshold for obtaining initial masks ($\tau_{b}$), and feature similarity threshold for merging ($\tau^\text{merge}_f$). We refer to the grid size parameter as `stride', where the stride is inversely proportional to the grid size. For instance, with a stride of 2 and a spatial dimension of 60$\times$60 of DINO feature embeddings, the grid size is reduced to 30$\times$30, resulting in 900 initial prompted masks (i.e., $K=900$). Conversely, a stride of 30 results in a grid size of 2$\times$2, yielding only 4 initial prompted masks (i.e., $K=4$).

In \cref{fig:effect-of-stride}, we present the performance of \methodName using strides of \{2, 4, 6, 8, 10, 15, 30\}. The stride significantly influences performance, with a stride of 4 yielding the best results, while larger strides such as 15 and 30 lead to diminished performance. This decline is attributed to the reduced number of masks generated at higher stride values.

We next examine the performance variation associated with different values of the bipartition threshold $\tau_{b}$, which is employed to binarize initial soft masks derived from prompting. A lower $\tau_{b}$ results in prompted masks having a larger area, while a higher $\tau_{b}$ leads to smaller mask areas. As illustrated in \cref{fig:effect-of-biparitition-tau}, optimal average precision and recall are achieved between 0.1 and 0.3, with the highest precision at $\tau_{b}=0.1$ and the highest recall at $\tau_{b}=0.3$. Performance in both metrics declines as $\tau_{b}$ increases, suggesting that \methodName benefits more from larger initial masks. Consequently, we set $\tau_{b}$ to 0.2 for all our experiments.

We also investigate the impact of the feature similarity threshold $\tau^\text{merge}_f$, which determines whether two masks should be merged. A lower $\tau^\text{merge}_f$ value leads to the merging of mask pairs even with minimal similarity, while a higher value requires a high degree of similarity for merging.
\cref{fig:effect-of-dino-merge-tau} demonstrates that our method's performance significantly varies with $\tau^\text{merge}_f$ values between 0.0 and 0.3, but levels off at higher values. This plateau suggests that beyond a $\tau^\text{merge}_f$ of 0.3, few mask pairs meet the merging criterion, thereby minimally impacting the overall performance metrics. Consequently, we opt for a $\tau^\text{merge}_f=0.1$  in our experiments.

\input{figures/hyperparameters}

%% file: tables/comparison-sota.tex
{
\setlength\tabcolsep{0.pt}
\begin{table*}[!t]
\centering
\fontsize{7.1pt}{9.5pt}\selectfont
\begin{tabular}{c|cc|cc|cc|cc|cc|cc|cc}
& \multicolumn{2}{c|}{COCO2017} & \multicolumn{2}{c|}{COCO-20K}& \multicolumn{2}{c|}{LVIS}& \multicolumn{2}{c|}{KITTI}& \multicolumn{2}{c|}{Objects365$^\dagger$}& \multicolumn{2}{c|}{SA-1B}& \multicolumn{2}{c}{Average} \\
method & $\maskap$ & $\maskar$ & $\maskap$ & $\maskar$& $\maskap$ & $\maskar$& $\maskap$ & $\maskar$& $\boxap$ & $\boxar$& $\maskap$ & $\maskar$& $\simpleap$ & $\text{AR}_{100}$\\ \shline
\multicolumn{15}{l}{\textit{Training-free methods}}\\ \hline
\multicolumn{1}{l|}{TokenCut~\cite{wang2022tokencut}} & 2.0 & 4.4 & 2.7 & 4.6 & 0.9 & 1.8 & \textbf{0.3} & 1.5 
& 1.1 & 2.1 & 1.0 & 0.3 & 1.3 & 2.5 \\
\multicolumn{1}{l|}{MaskCut~\cite{wang2023cvpr}} & 2.2 & 6.9 & \textbf{3.0} & 6.7 & 0.9 & 2.6 & 0.2 & \textbf{2.2} 
& 1.7 & 4.0 & 0.8 & 0.6 & 1.5 & 3.5\\
\multicolumn{1}{l|}{ProMerge} & \textbf{2.4} & \textbf{7.5} & \textbf{3.0} & \textbf{7.4} & \textbf{1.3} & \textbf{3.3} & \textbf{0.3} & 1.9 & \textbf{2.2} & \textbf{6.0} & \textbf{1.2} & \textbf{0.8} & \textbf{1.7}&\textbf{4.5}\\
\hline
\multicolumn{15}{l}{\textit{Models trained with pseudo-labels}}\\ \hline
\multicolumn{1}{l|}{CutLER$^\ddagger$~\cite{wang2023cvpr}} & 8.7 & 24.9 & 8.9 & 25.1 & 3.4 & 16.6 & 3.9 & 23.3 & 11.5 & 34.3 & 5.5 & 13.5 & 7.0 &22.9\\
\multicolumn{1}{l|}{ProMerge+} & \textbf{8.9} & \textbf{25.1} &  \textbf{9.0} & \textbf{25.3} & \textbf{4.0} & \textbf{17.7} & \textbf{5.4} & \textbf{25.7} & \textbf{12.2} & \textbf{35.8} & \textbf{7.8} & \textbf{16.3} & \textbf{7.9}& \textbf{24.3}
\end{tabular}
\caption{\textbf{Comparison between training-free methods (top) and models trained with pseudo-labels (bottom).} CutLER and ProMerge+ are trained with pseudo-labels generated by MaskCut and ProMerge, respectively. $^\dagger$Only ground-truth bbox annotations available.
$^\ddagger$Re-implemented with a single round of training for a fair comparison.
}
\label{tab:sota}
\end{table*}
}

%% file: tables/comparison-speed.tex
{
\begin{figure}[!t]
    \begin{minipage}[b]{0.4\textwidth}
    \fontsize{7.5pt}{10pt}\selectfont
        \centering
        \begin{tabular}{l|c}
        \multicolumn{1}{c|}{method} & FPS \\ \shline
        TokenCut~\cite{wang2022tokencut} & 0.34 \\
        MaskCut~\cite{wang2023cvpr} & 0.15 \\ \hline
        \methodName & \textbf{0.54}
        \end{tabular}
        \captionof{table}{\textbf{Speed comparison.} Our method is approximately 3.6 times faster in FPS compared to MaskCut.}
        \label{tab:speed-comparison}
    \end{minipage}
    \qquad
    \begin{minipage}[b]{0.505\linewidth}
        \centering
        \includegraphics[width=\textwidth]{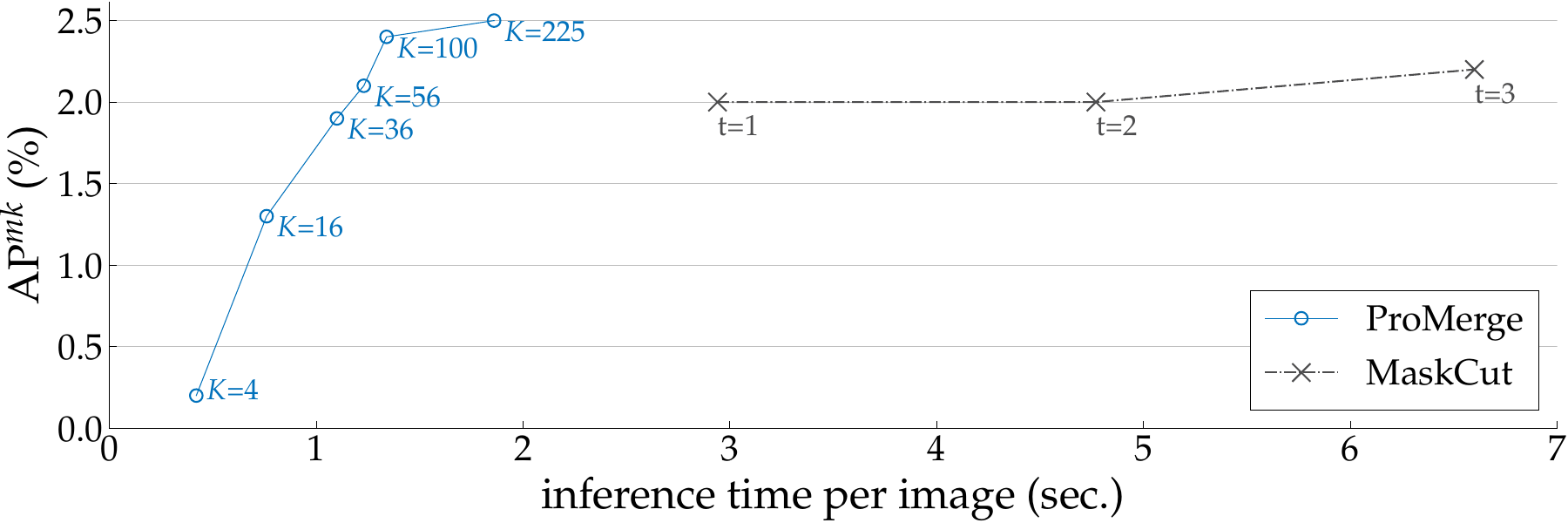}
        \vspace{-7mm}
        \caption{\textbf{$\maskap$ vs inference time on COCO2017.} $K$ denotes the number of prompt tokens. MaskCut uses $t$=3 by default in the original paper.}
        \label{fig:stride-speed}
    \end{minipage}
\end{figure}
}

%% file: tables/effect-of-each-component.tex
{
\setlength{\tabcolsep}{3pt}
\begin{table}[!t]
\fontsize{7.5pt}{10pt}\selectfont
\centering
\begin{tabular}{cccc|ccc}
Prompting & Merging & Mask Pruning & CC Splitting & $\maskapfifty$ & $\maskap$ & $\maskar$ \\ \shline
 \checkmark & \xmark & \xmark& \xmark & 0.8 & 0.5 & 1.6 \\
 \checkmark & \checkmark & \xmark& \xmark & 0.6 & 0.4 & 1.0 \\
 \checkmark & \checkmark & \checkmark & \xmark & 4.4 & 2.1 & 4.7 \\
 \checkmark & \checkmark & \checkmark  & \checkmark & \cellcolor{grey_cell}\textbf{5.6} & \cellcolor{grey_cell}\textbf{2.4} & \cellcolor{grey_cell}\textbf{7.5} 
\end{tabular}
\caption{\textbf{Effect of individual components.} A naive prompting and merging approach suffers poor performance, while applying the proposed background-based mask pruning (Mask Pruning) allows for a notable increase in performance, which is further enhanced by connected component splitting (CC Splitting).
Default settings are marked in \colorbox{grey_cell}{gray}.}
\label{tab:each-component}
\end{table}
}

%% file: tables/effect-of-voting.tex
{
\setlength{\tabcolsep}{0.25pt}
\begin{table}[!t]
\begin{subtable}[h]{0.45\textwidth}
\centering
\fontsize{7.5pt}{10pt}\selectfont
    \begin{tabular}{c|ccc|ccc}
    voting & $\boxapfifty$ & $\boxap$ & $\boxar$ & $\maskapfifty$ & $\maskap$ & $\maskar$ \\ \shline
    \xmark & 4.1 & 1.6 & 5.0 & 3.2 & 1.2 & 4.2 \\
    \checkmark & \cellcolor{grey_cell}\textbf{6.0} & \cellcolor{grey_cell}\textbf{3.0} & \cellcolor{grey_cell}\textbf{8.6} & \cellcolor{grey_cell}\textbf{5.6} & \cellcolor{grey_cell}\textbf{2.4} & \cellcolor{grey_cell}\textbf{7.5} \\
    \multicolumn{1}{c}{}&&&\multicolumn{1}{c}{}&&&\\
    \end{tabular}
    \caption{Effect of voting}
    \label{tab:effect-of-voting}
\end{subtable}%  % note '%' after \end{subtable}
\quad
\begin{subtable}[h]{0.45\textwidth}
\centering
\fontsize{7.5pt}{10pt}\selectfont
\begin{tabular}{c|ccc|ccc}
filter. method & $\boxapfifty$ & $\boxap$ & $\boxar$ & $\maskapfifty$ & $\maskap$ & $\maskar$ \\ \shline
IoA & 4.7 & 2.1 & 6.8 & 4.0 & 1.7 & 5.8   \\
feat. & 5.0 & 2.3 & 5.9 & 4.6 & 2.0 & 5.2  \\
Cascade & \cellcolor{grey_cell} \textbf{6.0} & \cellcolor{grey_cell} \textbf{3.0} & \cellcolor{grey_cell} \textbf{8.6} & \cellcolor{grey_cell} \textbf{5.6} & \cellcolor{grey_cell} \textbf{2.4} & \cellcolor{grey_cell} \textbf{7.5} \\
\end{tabular}
\caption{Impact of filtering criteria}
\label{tab:effect-of-filtering-conditions}
\end{subtable}
\caption{\textbf{Influence of the voting strategy and filtering conditions.} \textbf{(Left)} We note that using the proposed pixel-wise voting for obtaining a representative background mask allows for a notable gain in performance. \textbf{(Right)} Comparing the Intersection-over-Area-based filtering (denoted as IoA) and the feature similarity-based filtering (denoted as feat.), the former demonstrates higher recall, whereas the latter excels in precision.
Our proposed Cascade filtering outperforms both in all evaluated metrics, showcasing its effectiveness.
}
\end{table}
}

%% file: tables/effect-of-filtering-condition.tex
{
\setlength{\tabcolsep}{2.5pt}
\begin{table}[!t]
\fontsize{7.5pt}{10pt}\selectfont
\centering
\begin{tabular}{cc|ccc|ccc}
    feat. & IoA & $\boxapfifty$ & $\boxap$ & $\boxar$ & $\maskapfifty$ & $\maskap$ & $\maskar$ \\ \shline
    \xmark & \checkmark & 5.3 & 2.3 & 7.9 & 4.4 & 1.9 & 6.9 \\
    \checkmark & \xmark & 5.8 & 2.7 & 8.4 & 4.9 & 2.2 & 7.2 \\
    \checkmark & \checkmark & \cellcolor{grey_cell}\textbf{6.0}  & \cellcolor{grey_cell}\textbf{3.0}  & \cellcolor{grey_cell}\textbf{8.6} & \cellcolor{grey_cell}\textbf{5.6} & \cellcolor{grey_cell}\textbf{2.4} & \cellcolor{grey_cell}\textbf{7.5}
\end{tabular}
\caption{\textbf{Effect of merging conditions.} Considering both feature similarity (denoted as feat.) and intersection-over-area (denoted as IoA) for merging a pair of masks yields further gain.}
\label{tab:effect-of-merging-conditions}
\end{table}
}

%% file: figures/hyperparameters.tex
\begin{figure}[t]
     \centering
     \begin{subfigure}[b]{0.315\textwidth}
         \centering
         \includegraphics[width=\textwidth]{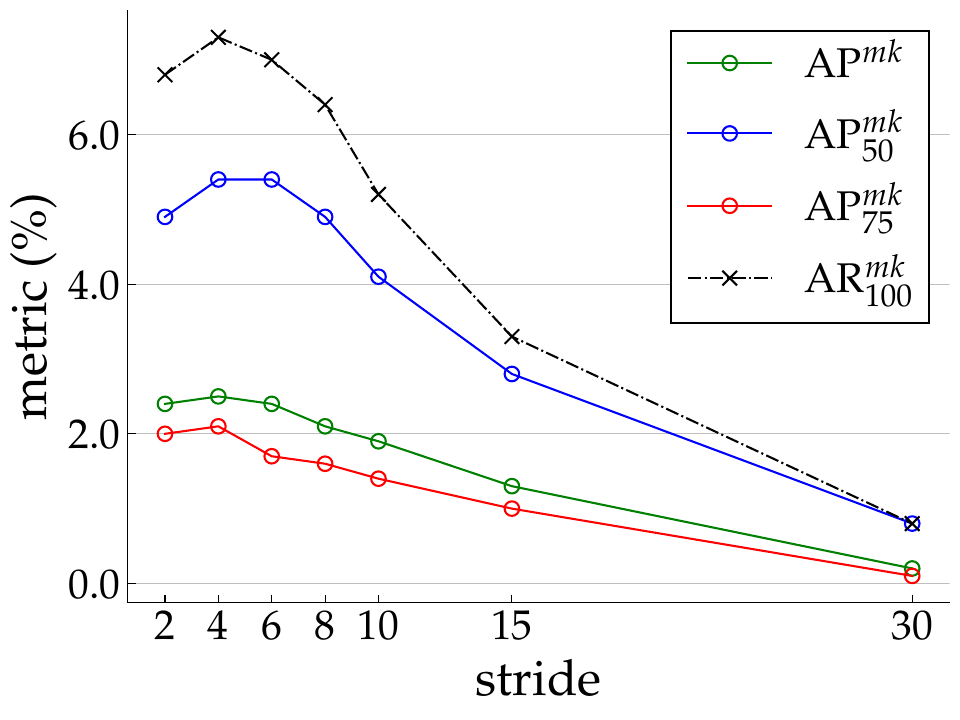}
         \caption{Effect of stride}
         \label{fig:effect-of-stride}
     \end{subfigure}
     \hfill
     \begin{subfigure}[b]{0.315\textwidth}
         \centering
         \includegraphics[width=\textwidth]{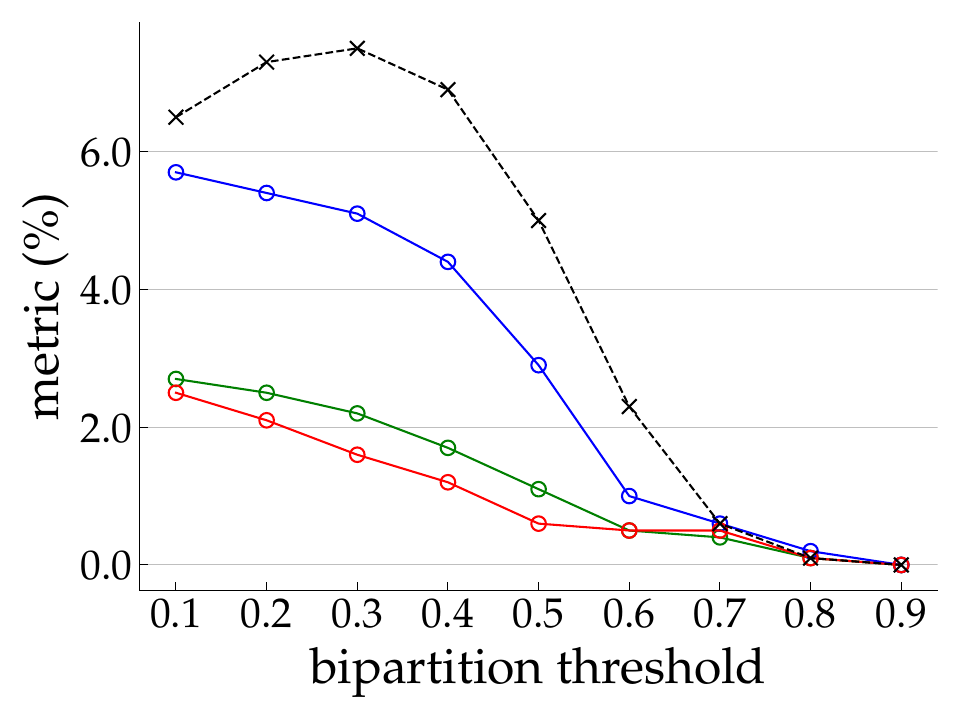}
         \caption{Effect of $\tau_b$}
         \label{fig:effect-of-biparitition-tau}
     \end{subfigure}
     \hfill
     \begin{subfigure}[b]{0.315\textwidth}
         \centering
         \includegraphics[width=\textwidth]{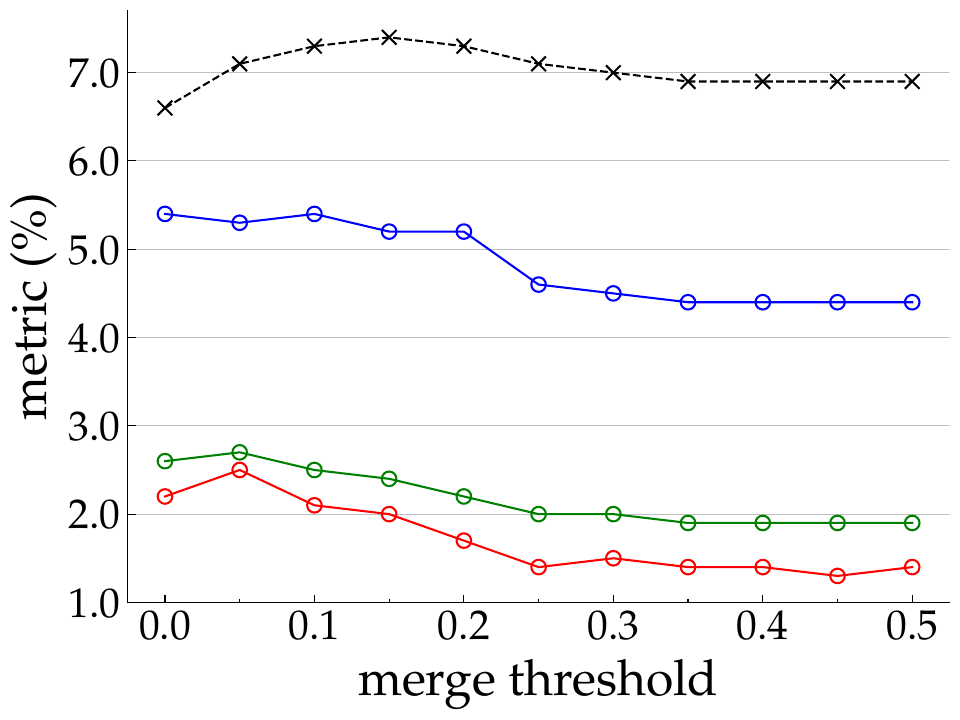}
         \caption{Effect of $\tau^\text{merge}_f$}
         \label{fig:effect-of-dino-merge-tau}
     \end{subfigure}
    \caption{\textbf{Hyperparameter analysis.} We note that stride (i.e., the Euclidean distance between two nearest point prompts in a regular 2D grid), bipartition threshold ($\tau_b$) for binarizing prompted masks, and feature similarity threshold ($\tau^\text{merge}_f$) for merging masks play important roles in our approach. The default setting uses stride=4, $\tau_b$=0.2, and $\tau^\text{merge}_f$=0.1, respectively. Best viewed in color.
    } 
    \label{fig:hyperparameters-analysis}
\end{figure}

%% file: sections/05-discussion.tex
\section{Discussion}
We observe that a simple approach of prompting and merging masks is competitive with computationally intensive normalized-cut-based approaches~\cite{wang2022tokencut,wang2023cvpr}. We primarily attribute the success of \methodName to two properties: (i) unlike in~\cite{wang2022tokencut,wang2023cvpr}, we do not assume a fixed number of mask predictions per image and allow the algorithm to flexibly make predictions, which we find particularly helpful when multiple objects are present in a given image;
(ii) we use the proposed sophisticated background-based filtering method, which excludes masks that overlap with the background of an image. 

\methodName shows promise but still has room to improve compared to fully-supervised methods, such as SAM~\cite{kirillov2023iccv}.
We believe that the gap partly results from utilizing self-supervised features that are not trained with a pretext task driven by object localization or segmentation.
Training and adopting self-supervised features specifically tailored for object localization or segmentation could significantly bridge this gap.

%% file: sections/06-conclusion.tex
\section{Conclusion}
Our work introduces Prompt and Merge (\methodName), a novel method to unsupervised instance segmentation that capitalizes on the strong local correspondences afforded by self-supervised visual features.
By iteratively merging initial patch groupings and employing a sophisticated background-based mask pruning technique, \methodName achieves competitive performance with a significant reduction in inference time compared to state\hyp{}of\hyp{}the\hyp{}art normalized\hyp{}cut\hyp{}based methods on standard benchmarks.
Moreover, the application of our method in training an object detector with pseudo-labels demonstrates superior performance, surpassing the leading unsupervised segmentation model. 

%% file: sections/acknowledgements.tex
\newpage
\mypara{Acknowledgements.}
This work was performed using resources provided by the Cambridge Service for Data Driven Discovery (CSD3) operated by the University of Cambridge Research Computing Service (www.csd3.cam.ac.uk), provided by Dell EMC and Intel using Tier-2 funding from the Engineering and Physical Sciences Research Council (capital grant EP/T022159/1), and DiRAC funding from the Science and Technology Facilities Council (www.dirac.ac.uk).
All authors appreciate the anonymous reviewers for the thoughtful suggestions and Fengting Yang for the kind advice on the rebuttal.
DL would like to also thank Eric Tran and Robert Wang for their support while DL conducted his research at Meta.
GS would like to thank Zheng Fang for the enormous support.

%% file: supplementary-sections/introduction.tex
We first provide pseudo-code for our approach (\cref{sec:pseudo-code})
along with further details about our experiments (\cref{sec:further-implementation-details}) and additional ablation studies (\cref{sec:further-ablation-studies}).
We also visualize qualitative results in \cref{sec:additional-visualizations}.

%% file: supplementary-sections/pseudo-code.tex
\section{Pseudo-code for \methodName}
\label{sec:pseudo-code}
In this section, we provide pseudo-code for our method (see \cref{alg:pseudo-code}) along with brief descriptions for four components: (i) prompting,
(ii) background aggregation,
(iii) Cascade filtering, and
(iv) merging. 
\input{algorithms/promerge}

\mypara{Prompting.}
Our approach generates initial mask proposals by employing point-prompting on visual features extracted from an image using an image encoder, such as a ViT~\cite{dosovitskiy2021iclr}, and creating affinity matrices based on cosine similarities between selected \textit{prompt tokens} and all patch tokens.
In this prompting stage, we use a hyperparameter for stride, representing the distance between two neighboring prompt tokens.
We use a default value of 4.
In our standard implementation, we use DINO~\cite{caron2021dino} features of spatial dimensions $60\times60$ and generate 225 mask proposals, which we then classify into background and foreground categories.

We use another hyperparameter, $\tau_b$, to threshold the cosine similarity between the embeddings of the seed prompt token and the other patches in the affinity matrix. This step allows us to translate continuous affinities into discrete masks for each seed prompt. For our experiments, we set $\tau_{b} = 0.2$, unless otherwise stated. 

Following the prompting, we split connected components from each foreground mask into separate masks, which helps separate a single large mask covering multiple instances. 

\mypara{Background aggregation.}
After prompting visual features, binary masks are classified as foreground or background.
Background masks are identified by the presence of numerous positive pixels along multiple image edges, and a representative background mask is created using a pixel-wise voting scheme from these candidates.

\mypara{Cascade filtering.}
In Cascade filtering, prompted masks are first sorted by area in ascending order, then processed sequentially to track new pixels added by each mask. Masks that significantly overlap with the background, as determined by Intersection over Area (IoA) and feature similarity, are excluded from the merging process.
For this, we use hyperparameters, $\tau_\text{IoA}^\text{bg}$ and $\tau_{f}^\text{bg}$.
If a new mask mainly introduces pixels that intersect with the background or shares a high feature similarity with the background, we do not include it in the merging process.
We set $\tau_\text{IoA}^\text{bg} = 0.8$ and $\tau_{f}^\text{bg} = 0.1$ by default.

\mypara{Merging.}
In our iterative clustering approach, filtered prompted masks are processed and merged in descending order of area, using the IoA metric and feature similarity.
Smaller masks merge with larger ones if their IoA area overlap exceeds $\tau_\text{IoA}^\text{merge}$ or if their similarity in feature space exceeds $\tau_{f}^\text{merge}$.
We set $\tau_\text{IoA}^\text{merge} = 0.1$ and $\tau_{f}^\text{merge} = 0.1$ for our experiments.
\newpage

%% file: algorithms/promerge.tex
\begin{algorithm}
\begin{algorithmic}
\caption{Pseudo-code of \methodName in a PyTorch-like style}\label{alg:pseudo-code}
\State \comment{\qcr{\# F: features (HxWxC)}}
\State \comment{\qcr{\# s: stride}}
\State \comment{\qcr{\# mm: matrix multiplication}}
\State \comment{\qcr{\# cc\textunderscore split: connected components splitting}}
\State \comment{\qcr{\# merge\textunderscore masks : merge clusters and proposal mask}}

\State
\State \comment{\qcr{\# Prompting}}
\State \qcr{bg\textunderscore masks, fg\textunderscore masks = [], []}
\State \qcr{for i in range(0, H, s):}
\State \qquad \qcr{for j in range(0, W, s):}
\State \qquad\qquad \qcr{prompt\textunderscore token = F[i, j, :]}
\State \qquad\qquad \qcr{mask = mm(F, prompt\textunderscore token)$>\tau_b$}
\State \qquad\qquad \qcr{if is\textunderscore background(mask):}
\State \qquad\qquad\qquad \qcr{bg\textunderscore masks.append(mask)}
\State \qquad\qquad \qcr{else:}
\State \qquad\qquad\qquad \qcr{fg\textunderscore masks.append(cc\textunderscore split(mask))}
\State
\State \comment{\qcr{\# Background aggregation}}

\State \qcr{for bg\textunderscore mask in bg\textunderscore masks:}
\State \qquad \qcr{voted\textunderscore bg += bg\textunderscore mask}
\State \qcr{voted\textunderscore bg = round(voted\textunderscore bg / len(bg\textunderscore masks))}  \comment{\qcr{\# pixel-wise voting}}

\State
\State \comment{\qcr{\# Cascade filtering}}
\State \qcr{filtered\textunderscore fg = []}
\State \qcr{sort(fg\textunderscore masks, x=\text{lambda} x:\text{sum}(x))}  \comment{\qcr{\# sort by size (asc.)}}
\State \qcr{fg\textunderscore seen = zeros(H,W)}
\State \qcr{bg\textunderscore ft = mean(F[voted\textunderscore bg, :], axis=0)}
\State \qcr{for fg in fg\textunderscore masks:}
\State \qquad \qcr{fg\textunderscore unseen = fg - fg[fg\textunderscore seen$>$0]}
\State \qquad \qcr{intersection = fg\textunderscore unseen \& voted\textunderscore bg}
\State \qquad \qcr{fg\textunderscore ft = mean(F[fg, :], axis=0)}
\State \qquad \qcr{if sum(intersection) / sum(fg\textunderscore unseen)$<\tau_\text{IoA}^\text{bg}$ and }
\State \qquad \qquad \qcr{(bg\textunderscore ft @ fg\textunderscore ft)$<\tau_{f}^\text{bg}$:}
\State \qquad \qquad \qcr{fg\textunderscore seen += fg\textunderscore unseen}
\State \qquad \qquad \qcr{filtered\textunderscore fg.append(fg)}

\State
\State \comment{\qcr{\# Merging}}
\State \qcr{clusters = set()}
\State \qcr{sort(filtered\textunderscore fg, x=\text{lambda} x:\text{sum}(x), reverse=True)}
\State \qcr{for fg in filtered\textunderscore fg:}
\State \qquad\qcr{masks\textunderscore to\textunderscore merge = []}
\State \qquad \qcr{for c in clusters:}
\State \qquad \qquad \qcr{c\textunderscore mean = mean(F[c, :], axis=0)}
\State \qquad \qquad \qcr{fg\textunderscore mean = mean(F[fg, :], axis=0)}
\State \qquad \qquad \qcr{if mm(fg\textunderscore mean, c\textunderscore mean)$>\tau_{f}^\text{merge}$:}
\State \qquad \qquad \qquad \qcr{masks\textunderscore to\textunderscore merge.append(c)}
\State \qquad \qquad \qcr{elif (sum(fg \& c) / sum(fg))$>\tau_\text{IoA}^\text{merge}$:}
\State \qquad \qquad \qquad \qcr{masks\textunderscore to\textunderscore merge.append(c)}
\State \qquad \qquad \qcr{if len(masks\textunderscore to\textunderscore merge) == 0:}
\State \qquad \qcr{if len(masks\textunderscore to\textunderscore merge) == 0:}
\State \qquad \qquad \qcr{clusters.add(fg)}
\State \qquad \qquad \qcr{continue}
\State \qquad \qcr{merged\textunderscore mask = merge\textunderscore masks(masks\textunderscore to\textunderscore merge, fg)}
\State \qquad \qcr{clusters.replace(masks\textunderscore to\textunderscore merge, merged\textunderscore mask)}

\State
\State \comment{\qcr{\# Postprocessing}}
\State \qcr{for mask in clusters:}
\State \qquad \qcr{mask = dense\textunderscore crf(mask)}

\State \qcr{return clusters}
\end{algorithmic}
\end{algorithm}

%% file: supplementary-sections/further-implementation-details.tex
\section{Further implementation details}
\label{sec:further-implementation-details}

Here, we provide additional details about the datasets used in our paper and training of ProMerge+.

\mypara{Datasets.}
We evaluate our methods on six benchmarks, including COCO2017~\cite{lin2014eccv}, COCO20K~\cite{vo2020eccv}, LVIS~\cite{gupta2019cvpr}, KITTI~\cite{geiger2012cvpr}, subsets of Objects365~\cite{shao2019iccv} and SA-1B~\cite{kirillov2023iccv}.

COCO2017 and COCO20K are the standard datasets for object detection and segmentation.
COCO2017 is composed of 118K and 5K images for training and validation splits respectively, while COCO20K is composed of 20K images.
For all results on COCO2017, we use the 5K images in the validation split.

LVIS is a more challenging dataset for object detection and segmentation, with densely-annotated instance masks.
We test our performance on the validation set, which contains 245K instances on 20K images.
For KITTI and Objects365, we evaluate on 7K images following~\cite{wang2023cvpr} and a subset of 44K images in the val split, respectively. Lastly, for SA-1B, we assess on a subset of 11K images, which come with 100+ annotations per image on average.\footnote{The subset for SA-1B can be downloaded from \url{https://ai.meta.com/datasets/segment-anything-downloads/}}

\mypara{Training details of ProMerge+.}
For training ProMerge+, we follow the same training protocol as described in CutLER~\cite{wang2023cvpr}, and compare its performance with CutLER after a single training cycle.
For a fair comparison, we reimplement a single round training of CutLER using the official codebase. 
Specifically, we use Cascade Mask-RCNN~\cite{cai2018cvpr} and initialize the image encoder (i.e., ResNet-50 backbone~\cite{he2016cvpr}) with DINO pretrained weights~\cite{caron2021dino}.
We also leverage the copy-and-paste augmentation \cite{Ghiasi_2021_CVPR}.
We train the detector on ImageNet~\cite{deng2009cvpr} images with their pseudo-labels obtained via \methodName, after removing noisy masks whose area is smaller than 5\% of the size of the corresponding input image. For training, we use a base learning rate of $0.005$ for 80K steps that drops to $0.001$ for the remaining 80K iterations and use a weight decay of $5e^{-5}$. The training is based on the Detectron2 framework~\cite{wu2019detectron2}.
\newpage

%% file: supplementary-sections/further-ablation-studies.tex
\section{Further ablation studies}
\label{sec:further-ablation-studies}
In this section, we conduct further ablation studies regarding the proposed method.

\mypara{Difference between Cascade filtering and non-maximum suppression.}
We note that, at a glance, the proposed Cascade filtering (CF) approach can bear resemblance to the commonly used non-maximum suppression (NMS). However, there are crucial differences: CF
(i) filters mask proposals by comparing them to a background mask, whereas NMS does so by comparing the proposals to each other;
(ii) considers the new pixel regions that are not part of any previously accepted proposals combined; and
(iii) takes into account feature similarity with the background as well as pixel overlap.
Indeed, we observe the stark difference in $\maskap$---2.4\% for CF vs 0.8\% for NMS on COCO2017, showing the importance of pruning noisy background masks via CF. 

\input{tables/prompting-method}
\mypara{Prompting methods.} In our paper, we use features equally spaced in a regular 2D grid as prompt tokens to obtain initial mask proposals. Here, we explore alternative prompting methods including random, iterative, attentive, and multi-head attention (MHA) prompting.

Random prompting selects patches randomly from all of the patch tokens (extracted from the image encoder) and uses them as prompt tokens.
Iterative prompting uses the initial grid of prompts, but shifts the prompt center over multiple iterations.
This prompting method first takes the prompt token from a regular grid, and finds the spatial center of the tokens in the initial mask proposal.
The token at the mask center is then selected as the new prompt token.
This process is repeated three times to find the optimal set of prompt tokens, with the goal of seeking prompts that represent the central components of objects. 
Attentive prompting identifies distinctive patch tokens, which serve as prompt tokens, by using a mode-seeking clustering algorithm (FINCH~\cite{finch}).
MHA prompting leverages the observation from \cite{caron2021dino} that the last self-attention layer of the DINO-ViT groups foreground objects. We first compute cosine similarities between the [CLS] token and query features from each head in the last self-attention layer of the DINO-ViT, producing multiple affinity maps. We then sum all the affinity maps and identify 2D coordinates whose cumulative affinities are within the top 5\%.  We then use these corresponding patch tokens as prompt tokens, and run inference without additional modifications to the \methodName pipeline.

As shown in~\cref{tab:prompt-type}, random prompting performs best among these alternative methods, showing notably higher recall than iterative, attentive, and MHA prompting. We conjecture that random prompting tends to cover diverse regions of an image, thus lifting recall. However, the default regular grid prompting that we employ in \methodName shows slightly higher recall than random prompting, as it is guaranteed to cover the entire image area, provided that the prompt tokens are sufficiently dense.

\newpage

%% file: tables/prompting-method.tex
\setlength\tabcolsep{1.5pt}
\begin{table}[!t]
\centering
\footnotesize
\begin{tabular}{cc|cc|cc|cc|cc}
\multicolumn{2}{c|}{regular grid} &
\multicolumn{2}{c|}{random} & \multicolumn{2}{c|}{iterative} &
\multicolumn{2}{c|}{attentive}&
\multicolumn{2}{c}{MHA}\\
$\maskap$ & $\maskar$ & $\maskap$ & $\maskar$ & $\maskap$ & $\maskar$ & $\maskap$ & $\maskar$ & $\maskap$ & $\maskar$\\ \shline
\cellcolor{grey_cell}\textbf{2.4} & \cellcolor{grey_cell}\textbf{7.5} & \textbf{2.4} & 7.3 & \textbf{2.4} & 5.3 & 2.0 & 5.6 & 2.2 & 6.3
\end{tabular}
\caption{\textbf{Effect of prompting methods.} The regular grid prompting is used by default in our method. Default setting is marked in \colorbox{grey_cell}{gray}.}
\vspace{-4mm}
\label{tab:prompt-type}
\end{table}

%% file: supplementary-sections/further-visualisations.tex
\section{Further visualizations}
\label{sec:additional-visualizations}
\input{figures/visual-comparison}
Here, we first showcase qualitative examples of training-free methods including \methodName, TokenCut~\cite{wang2022tokencut}, and MaskCut~\cite{wang2023cvpr}. 
Then, we visualize successful and failure cases of our approach.

\subsection{Qualitative comparison}
In~\cref{fig:visual-comparison}, we can see that both TokenCut and MaskCut struggle with segmenting multiple instances in an image due to their reliance on a predefined number of predictions per image (set to 3 in the original paper), whereas our approach flexibly segments numerous objects according to the input image.

\input{figures/success-cases}

\subsection{Success and failure cases of \methodName}
We show successful cases with multiple instance masks per image in \cref{fig:successful-cases}.
In \cref{fig:failure-cases}, on the other hand, we visualize typical failure cases in which \methodName undersegments multiple neighboring instances or oversegments an instance due to occlusion.
We attribute these artifacts to using visual features that are not explicitly trained for grouping pixels of an object based on an underlying semantic understanding.
That is, \methodName is not aware of the semantic boundaries of a single instance and is thus inclined to make mistakes in regions where multiple objects of the same category, sharing the same color or texture, are adjacent, or different parts of an object are located remotely.
\newpage

\input{figures/failure-cases}

\newpage

%% file: figures/visual-comparison.tex
\begin{figure}[t]
    \centering
    \includegraphics[width=.98\textwidth]{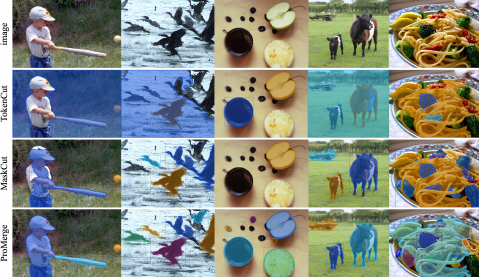}
    \caption{\textbf{Qualitative comparison between training-free unsupervised methods.}
    TokenCut~\cite{wang2022tokencut} and MaskCut~\cite{wang2023cvpr} fail to appropriately segment multiple instances. In contrast, \methodName (ours) successfully identifies multiple objects in an image. Zoom in for detail.}
    \label{fig:visual-comparison}
\end{figure}

%% file: figures/success-cases.tex
\newpage
\begin{figure}[!t]
    \centering
    \includegraphics[width=0.98 \textwidth]{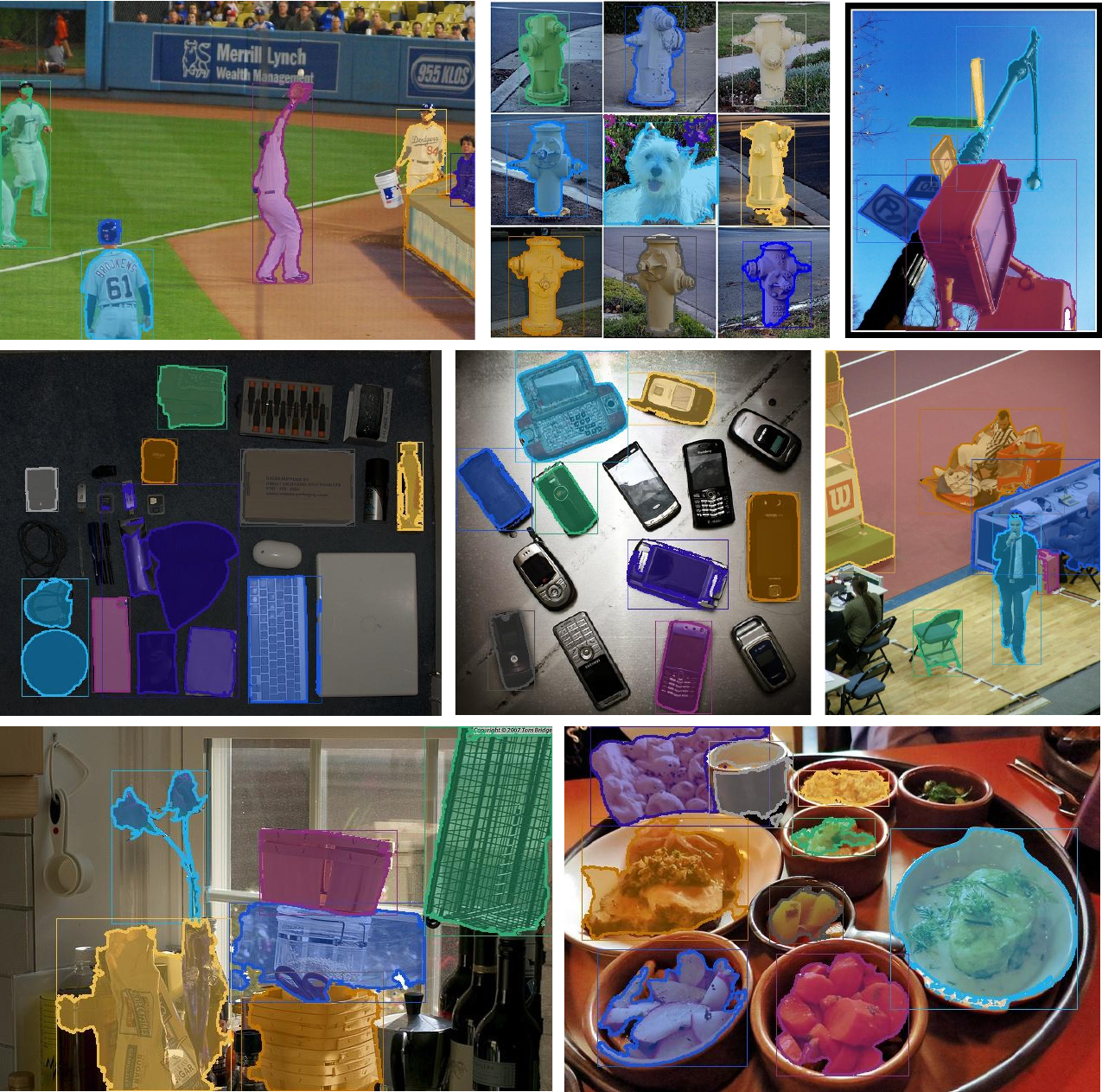}
    \vspace{-2mm}
    \caption{\textbf{Successful cases of ProMerge.} We provide additional visualizations that highlight ProMerge's ability to segment multiple distinct objects per image.}
    \label{fig:successful-cases}
\end{figure}
\newpage

%% file: figures/failure-cases.tex
\begin{figure}[H]
    \centering
    \includegraphics[width=0.98\textwidth]{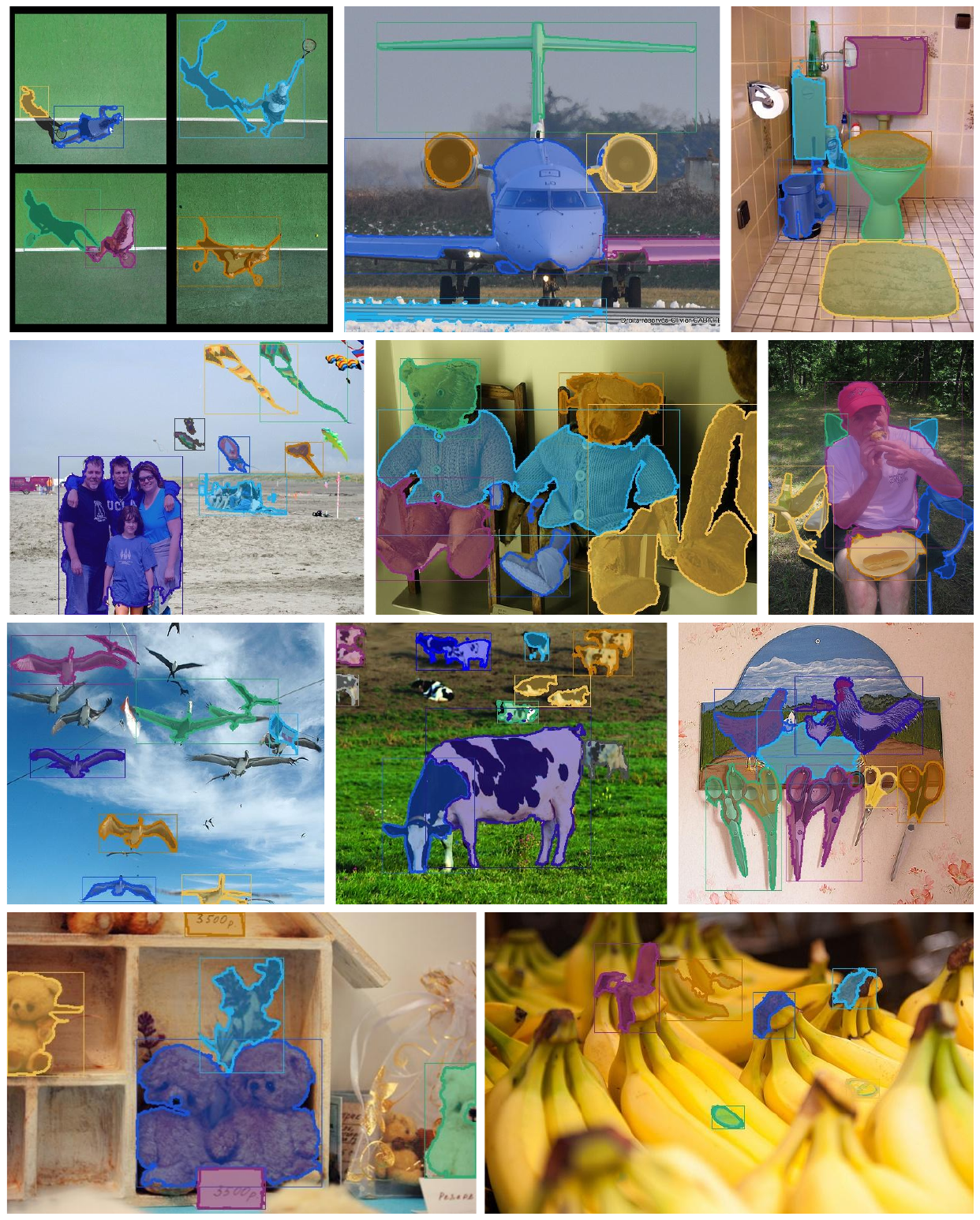}
    \vspace{-8mm}
    \caption{\textbf{Typical failure cases of ProMerge.} As ProMerge does not use features with an explicit understanding of a concept (i.e., class), ProMerge predicts lower quality masks for multiple instances of the same concept that are densely packed. ProMerge also struggles with recognizing a single object whose parts are scattered across different image regions due to occlusion.
    For example, adjacent teddy bears with similar textures are segmented as one (bottom left).
    In the case in which a chair is occluded by a person sitting, 
    ProMerge generates four separate annotations for different parts of the same chair (right image in the second row).
    }
    \label{fig:failure-cases}
\end{figure}